\documentclass{esannV2}
\usepackage{graphicx}
\usepackage[latin1]{inputenc}
\usepackage{amssymb,amsmath,array}
\usepackage{caption}
\usepackage{subcaption}
\usepackage{color}
\usepackage{multicol}

\begin{document}
\title{Sliced-Wasserstein normalizing flows:\\ beyond maximum likelihood training}

\author{Florentin Coeurdoux$^1$, Nicolas Dobigeon$^{1,2}$ Pierre Chainais$^3$%
%
\thanks{This work was supported by the Artificial Natural Intelligence Toulouse Institute (ANITI, ANR-19-PI3A-0004), the AI Sherlock Chair (ANR-20-CHIA-0031-01), the ULNE national future investment programme (ANR-16-IDEX-0004) and the Hauts-de-France Region.}
%
\vspace{.3cm}\\
%
1- University of Toulouse, IRIT/INP-ENSEEIHT, F-31071 Toulouse, France
%
\vspace{.1cm}\\
2- Institut Universitaire de France (IUF), France
\vspace{.1cm}\\
3- Univ. Lille, CNRS, Centrale Lille, UMR 9189 CRIStAL, F-59000 Lille, France
}

\maketitle

\begin{abstract}
Despite their advantages, normalizing flows generally suffer from several shortcomings 
including their tendency to generate unrealistic data (e.g., images) and their failing to detect out-of-distribution data.
One reason for these deficiencies lies in the training strategy which traditionally exploits a maximum likelihood principle only. This paper proposes a new training paradigm based on a hybrid objective function combining the maximum likelihood principle (MLE) and a sliced-Wasserstein distance. Results obtained on synthetic toy examples and real image data sets show better generative abilities in terms of both likelihood and visual aspects of the generated samples. Reciprocally, the proposed approach leads to a lower likelihood of out-of-distribution data, demonstrating a greater data fidelity of the resulting flows.
\end{abstract}

\section{Introduction}
Approximating probability distributions thanks to normalizing flows (NFs) has proven to be a powerful approach to accurately represent the underlying processes at the origin of collected data. NFs are designed to provide a tractable approximation of the data-generating density $p_{X}$ by transforming a base normal distribution through a series of bijective transformations \cite{papamakarios2021normalizing}. The usual approach to train such architectures relies on the principle of maximum likelihood estimation (MLE), i.e., maximizing  the joint density of the observed data with respect to (w.r.t.) the parameters of the network.
The constraint of relying on a parametric family of distributions may however raise crucial issues. Indeed, the optimality of MLE holds only when there is no model misspecification, i.e., the true data distribution $p_{X}$ belongs to the family that can be represented by  the optimized model. In practice, it is difficult to ensure a priori that the chosen family of functions is able to accurately model the targeted distribution. Hence the choice of the learning objective becomes largely an essential but most often empirical question. 
Moreover, training NF explicitly uses a Gaussian likelihood, i.e,  a function of the two first moments. Hence, optimizing a Gaussian likelihood function leaves any higher order moments completely free. A more refined way of fully characterizing the targeted distribution would be to match all the other higher order moments as well. Clearly, using a statistical distance between distributions during the training process would shift the optimization task from a nonlinear regression problem w.r.t. the likelihood parameters to a more relevant problem of looking for the best matching between the generated distribution and the targeted one. %
Grover \emph{et al.} \cite{grover2018flowgan} proposed a hybrid objective that bridges implicit and prescribed learning by combining MLE and adversarial training using a GAN. The hybrid objective has a balancing effect between perceptually good-looking samples and an accurate density estimation of the inputs. The authors also demonstrate that this hybrid objective has a regularizing effect, which permits the model to outperform MLE as well as adversarial learning. However the choice of using an adversarial architecture is accompanied by the well-documented drawbacks of GANs. An adversarial architecture requires the training of an additional discriminator which is notoriously unstable, can lead to mode collapse \cite{salimans2016improved} and  can produce overconfident predictions from out-of-distribution (OoD) inputs \cite{hendrycks17baseline}. 

To overcome the issues mentioned above, this paper introduces a novel hybrid loss function to be used to train NF.  As suggested above, in addition to the conventional MLE-based term, the proposed hybrid training loss also incorporates a term to  measure the discrepancy between the generated and the targeted distribution. This term derives from the Sliced-Wasserstein distance (SW) \cite{rpdb12sw} between the true data  distribution and the generated samples. Experimental results show that augmenting the MLE objective with this term consistently achieves higher likelihood as well as better quality of the generated samples. It also demonstrates better OoD detection capabilities compared to classical training of flow-based models. 
%
Section 2 introduces the proposed method referred to as sliced-Wasserstein NF (SW-NF) and its hybrid learning objective. Section 3 illustrates its performances on numerical experiments. Conclusions and prospects are reported in Section 4. 

\section{Sliced-Wasserstein flows}\label{sliced-wasserstein-flows}

\subsection{Normalizing flows}

NFs are a flexible class of deep generative networks
that learn a change of variable between two probability
distributions $p_{X}$ and $p_{Z}$ through an invertible
transformation
$f_{\theta}: X \mapsto Z=f_{\theta}(X)$
parametrized by $\theta$ \cite{papamakarios2021normalizing}. In general, 
$p_X$ is only known through samples
$x=\left\{x_n\right\}_{n=1}^N$ and, for tractability purpose, $p_Z$ is chosen as a centered normal distribution
with unit variance. The parameters $\theta$ defining the
operator $f_{\theta}$ are then adjusted according the MLE principle and exploiting 
the change of variable 
\begin{equation}
p_{X}(x)=p_{Z}\left(f_{\theta}(x)\right)\left|\operatorname{det} J_{f_{\theta}^{-1}}\right| \quad \textrm{with} \quad J_{f_{\theta}^{-1}}=\frac{\partial f_{\theta}^{-1}}{\partial x}
\label{change_var}
\end{equation}
In other words, the network is trained by minimizing the negative log-likelihood (NLL) or equivalently the loss function denoted by $\mathcal{L}_{\text {MLE}}(x; \theta) = - \log(p_{X}(x))$.
Without loss of generality, this work focuses on NFs based on affine coupling layers. Examples of such flows include RealNVP \cite{dinh2017realnvp}, Glow \cite{kingma2018glow} among others. 


\subsection{Sliced-Wassertein distance}

In recent years, Wasserstein distance, which is intimately related to the theory of optimal transport (OT), has received a considerable attention from the machine learning (ML) community because of its theoretical properties when comparing distribution. However, it suffers from strong computational and statistical limitations, which have severely hindered its effective use in problems in high dimensions. 
Several workarounds have been proposed to alleviate these issues and to enable the use of OT in ML applications. In particular, the Sliced-Wasserstein (SW) distance is an alternative OT metric \cite{rpdb12sw}. It has been increasingly popular since it benefits from a significantly reduced computational cost over the Wasserstein distance, especially on large-scale problems. In a nutshell, the SW distance compares high-dimensional distributions by comparing their projected 1d-distributions for which the computation of the Wasserstein distance is closed-form. According to a Monte Carlo principle the SW distance between two distributions $p_X$ and $p_Z$ empirically represented by two sets of samples $x$ and $z$, respectively, can be approximated by \emph{i)} drawing a large set of vectors $u_1,\ldots,u_J$  uniformly distributed over the unit sphere then $\emph{ii)}$  averaging the true 1d-Wasserstein distances between the slices of the two distributions along directions $u_i$. It will be denoted as $\mathcal{L}_{\text {SW}}(x, z)$ in what follows. Its formulation through its projections onto the unit sphere is well adapted when samples are vectors. 
%
%
%
Introduced by Nguyen et \emph{al.} \cite{nguyen2022convslicedw}, the so-called convolution SW (CSW) generalizes SW to images using a series of convolutions in the spirit of a multiresolution approach. We denote  $\mathcal{L}_{\text {CSW}}(x, z)$ the corresponding distance measure.

\subsection{Hybrid objective function}
The proposed SW-NF method builds on a NF neural architecture $f_\theta$ to target a normal latent distribution $p_Z$ so that the likelihood of the observed data $p_X$ is well-defined and tractable for exact evaluation and MLE training. 
Departing from conventional strategies deployed to train NFs, this work proposes to derive a hybrid objective function that binds the likelihood of a prescribed model to high order moment matching. To this aim the conventional MLE-based objective is augmented with an additional term measuring the discrepancy between the respective distributions of the original data $x \sim P_X$ and the generated data $\tilde{x} = f^{-1}_\theta(z)$. Note that the likelihood loss is prescribed on the latent space while the SW-based distance  between the generated and target distributions can be prescribed over the data space. Thus the proposed  hybrid objective is a combination of reconstruction and feature losses defined as
\begin{equation}
    \mathcal{L}(x,z;\theta)=\mathcal{L}_{\star \text{W}}(x, f^{-1}_\theta(z))+\alpha \mathcal{L}_{\text{MLE}}(x; \theta)
    \label{eq:loss_function}
\end{equation}
where $\alpha$ is a hyperparameter balancing the two terms and $\mathcal{L}_{\star\text {W}}$ refers to either $\mathcal{L}_{\text {SW}}$ for vector data sets or to $\mathcal{L}_{\text {CSW}}$ for image inputs, respectively. It is worth noting that the new objective function can be interpreted as a regularized counterpart of the change of variable on the data space. Moreover it has the great advantage of not depending on an auxiliary network as in \cite{grover2018flowgan}. Note that the SW-based discrepancy measure between the generative model and the data distributions can also be prescribed over the latent space by replacing the SW-based term in \eqref{eq:loss_function} by $L_{\star \text {W}}(z, f_\theta(x))$.



\section{Numerical experiments}

This section assesses the versatility and the accuracy of proposed SW-NF method through numerical experiments. First, experiments conducted on the Circle data set from scikit-learn are presented to provide some insights about key ingredients of the proposed approach. Then the performance of SW-NF is illustrated through  more realistic experimental settings exploiting the CIFAR-10 and SVHN image data sets. It is compared to the alternative training strategies which consist in relying on the sole MLE or SW terms in \eqref{eq:loss_function} and to the Flow-GAN method which hybridizes MLE and GAN losses \cite{grover2018flowgan}.
For all results reported below, the stochastic gradient descent is implemented in Pytorch, with the Adam optimizer, a learning rate of $10^{-4}$ and a batch size of $4096$ or $8192$ samples. When dealing with the toy example, the NF implementing the unknown mapping $f$ is chosen as a RealNVP  \cite{dinh2017realnvp} and the conventional SW distance is chosen for hybridization. When dealing with the image-driven experiments, the network architecture is Glow \cite{kingma2018glow} and the CSW is considered as a statistical distance. 
The proposed learning strategy is compared with conventional method from two task-driven perspectives, namely goodness-of-fit and OoD detection.

\begin{table}
    \centering
    \begin{tabular}{l|c|c|c|c|c}
    Objective & NLL & SW & $\left \| \kappa_{3} \right \|^2_2$ & $\left \| \kappa_{4} \right \|^2_2$\\ \hline
    MLE & 0.52 & 0.0033 & 0.2233 & 5.6124\\
    SW & 1.78 & $\mathbf{0.0007}$ & $\mathbf{0.0026}$ & $\mathbf{0.1822}$\\
    SW-Flow & $\mathbf{0.41}$ & 0.0008 & 0.0501 & 0.2259 \\
    Flow-GAN \cite{grover2018flowgan}& 0.51 & 1.23 & 0.4756 & 7.7725 
    \end{tabular}
    \caption{Circle data set: assessment of goodness-of-fit (for all metrics, the lower the better).\label{tab:circles}}
\end{table}

\begin{table}
    \centering
    \begin{tabular}{l|c|c|c|c|c}
    Objective & Inception & NLL (bits/dim) & CSW  & $\left \| \kappa_{3} \right \|^2_2$ & $\left \| \kappa_{4} \right \|^2_2$\\ \hline
    MLE & 2.42 & 3.54 & 1514.26 & 64.94 & 2462.37 \\
    SW & 1.28 & 9.81 & 1190.11 & 13.13 & \textbf{598.54} \\
    SW-Flow & 3.04 & \textbf{3.19} & \textbf{1014.26} & \textbf{6.24} & 656.37 \\
    Flow-GAN & \textbf{3.21} & 4.21 & 1621.78 & 72.3 & 3079.12 
    \end{tabular}
    \caption{CIFAR-10 data set: assessment of goodness-of-fit (for inception score, the higher the better; for all other metrics, the lower the better). \label{tab:cifar}}
\end{table}

\noindent{\textbf{Goodness-of-fit:} } We first study the goodness-of-fit of the targeted latent space through the learned inverse transform. This evaluation is conducted by evaluating not only the NLL but also the (C)SW distance and the $3$rd- and $4$th-order cumulants $\kappa_{3}$ and $\kappa_{4}$ which are expected to be equal to zero for the prescribed normal distribution $p_Z$. Table~\ref{tab:circles} and Table~\ref{tab:cifar} report the results reached by the proposed SW-NF approach for the circle-shaped and the CIFAR-10 data sets, respectively. Table~\ref{tab:cifar} also gives the inception score for visual inspection of the images.
Interestingly, the proposed SW-NF method provides significantly better NLL scores than the sole MLE-based learning strategy. For the two data sets, it also leads to competitive results in terms of (C)SW and normality features, reaching scores close to SW-based learning. In addition, when considering the CIFAR-10 images, it leads to higher inception score, thus suggesting higher visual quality of the generated samples.

\begin{table}[h!]
  \centering
  \begin{tabular}{l|c|c}
   Method & Moons & Blobs \\ \hline
   MLE & 0.37 & 0.54  \\
   SW & \textbf{0.61} & 0.70 \\
   SW-NF & 0.53 & \textbf{0.73} \\
   \end{tabular}
\caption{Circle data set: performance of OoD in term of AUROC (the closer to 1 the better).   \label{tab:auroc}}
\end{table}

\begin{figure}
     \centering
     \begin{subfigure}[b]{0.48\textwidth}
         \centering
         \includegraphics[width=\textwidth]{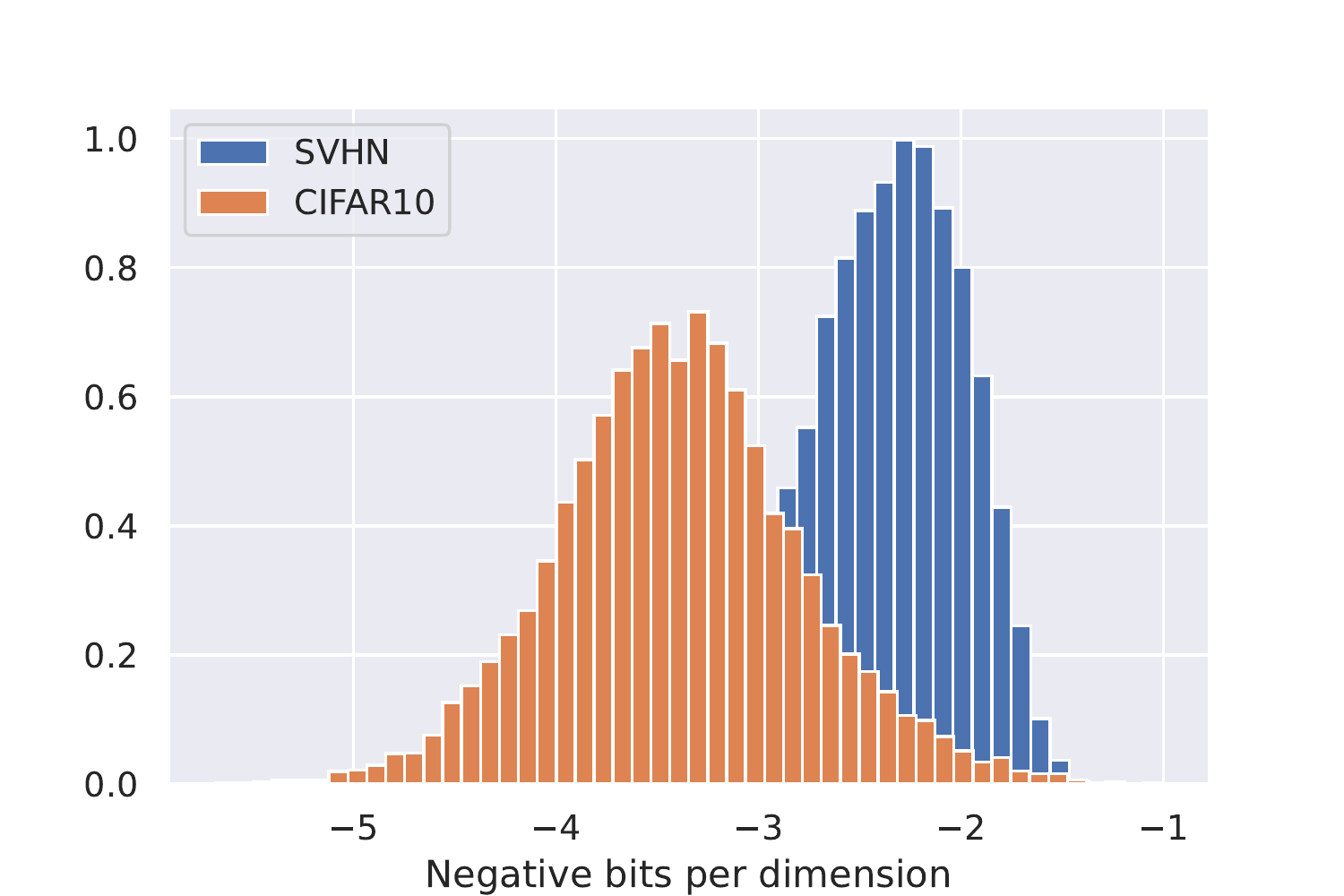}
         \caption{MLE training}
         \label{fig:gen-triangles}
     \end{subfigure}
     \hfill
     \begin{subfigure}[b]{0.48\textwidth}
         \centering
         \includegraphics[width=\textwidth]{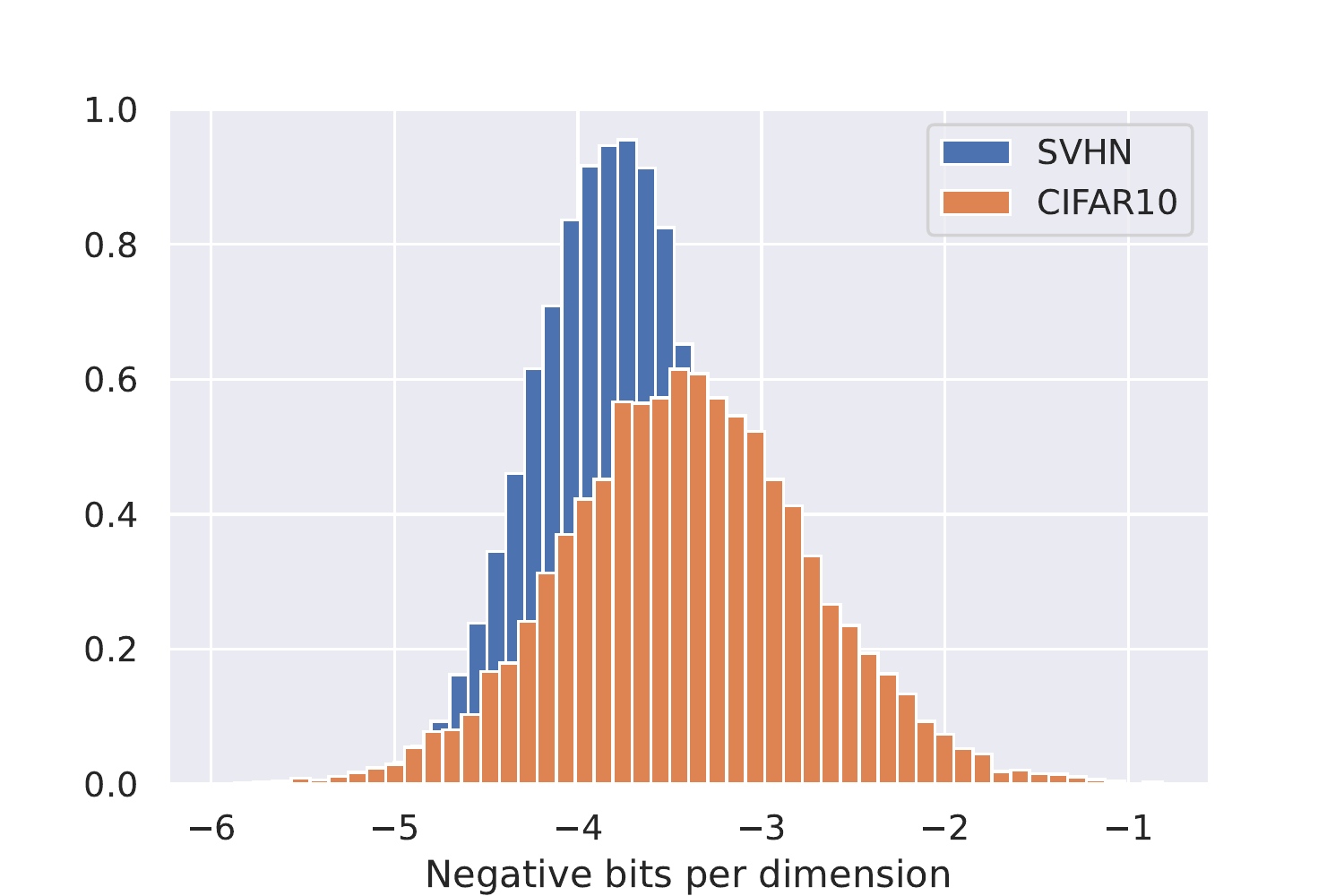}
         \caption{Hybrid training}
         \label{fig:gen-circles}
     \end{subfigure}
    \caption{Likelihood histogram for 1000 CIFAR-10 images (orange) and 1000 SVHN images (blue) prescribed by a CSW-NF trained on CIFAR-10 images. \label{fig:cifar-svhn}}
\end{figure}

\noindent{\textbf{Out-of-distribution detection:} } The second set of experiments assesses the ability of detecting  OoD data. Table~\ref{tab:auroc} reports the area under the receiving operator caracteristics (AUROC) for moon-shaped and blob-shaped data sets given a model trained on circles. In the absence of SW term in the training loss, the model constantly shows lower ability to discriminate OoD data from the training distribution data. In the context of CIFAR-10 data sets, the experimental setup of \cite{nalisnick2019generativedontknow} has been considered. Glow-based NFs have been trained on CIFAR-10 and we monitor the prescribed likelihood for both CIFAR-10 and SVHN images. Fig.~\ref{fig:cifar-svhn} (a) shows the results obtained by a sole MLE-based training: this model predicts a higher likelihood of OoD data. Fig.~\ref{fig:cifar-svhn} (b) depicts similar plots when considering the proposed hybrid loss: it leads to lower likelihood to OoD data coming from the SVHN data set.  

\section{Conclusion}

This paper introduces a new paradigm to train normalizing flow. It consists in augmenting the loss term derived from the conventional maximum likelihood principle with a discrepancy measure between the generated and targeted distributions. The resulting hybrid loss function thus combines a Gaussian likelihood with a (convolutional) sliced-Wasserstein distance between distributions. 
Numerical experiments show the better performance of the proposed hybrid training procedure in terms of perceptual as well as statistical quantitative metrics. 
On top of that, one observes a better robustness of the out-of-distribution behavior.
This works consists of a step towards the design of more powerful NF implemented as true generative models, beyond their simple use as nonlinear regressors structurally imposed by a conventional MLE training.
%


\renewcommand{\baselinestretch}{0.75}
\bibliographystyle{unsrt}

\bibliography{reference.bib}
\end{document}